\documentclass[sn-mathphys]{sn-jnl}

\jyear{2021}

\theoremstyle{thmstyleone}%

\theoremstyle{thmstyletwo}%

\theoremstyle{thmstylethree}%

\usepackage{graphicx}
\usepackage{caption}
\usepackage{subcaption}
\usepackage{longtable} 
\usepackage{enumitem} 
\usepackage{array}
\usepackage{gensymb} 
\usepackage{textcomp} 
\usepackage{pifont} 
\usepackage{numprint}
\usepackage{amsmath,amssymb,amsfonts}
\usepackage{adjustbox}
\usepackage{listings}
\usepackage{fixme}
\usepackage{hyperref}
\usepackage{float}
\fxsetup{status=draft}
\usepackage{amsmath,amsthm,amssymb,scrextend}
\usepackage{algorithm}

\begin{document}

\title[Image embedding]{Image Embedding for Denoising Generative Models}

\newcommand{\E}{\mathbb{E}}

\author*[1]{\fnm{Andrea} \sur{Asperti}}\email{andrea.asperti@unibo.it}

\author[2]{\fnm{Davide} \sur{Evangelista}}\email{davide.evangelista5@unibo.it}

\author[1]{\fnm{Samuele} \sur{Marro}}\email{samuele.marro@unibo.it}

\author[1]{\fnm{Fabio} \sur{Merizzi}}\email{fabio.merizzi@unibo.it}

\affil*[1]{\orgdiv{Department of Informatics: Science and Engineering (DISI)}, \orgname{University of Bologna}, \orgaddress{\street{Mura Anteo Zamboni 7}, \city{Bologna}, \postcode{40126},
\country{Italy}
}}

\affil*[2]{\orgdiv{Department of Mathematics}, \orgname{University of Bologna}, \orgaddress{\street{Piazza di Porta San Donato 1}, \city{Bologna}, \postcode{40126},
\country{Italy}
}}

\abstract{Denoising Diffusion models are gaining increasing popularity in the field of generative modeling for several reasons, including the simple and stable training, the excellent generative quality, and the solid probabilistic foundation. In this article, we address the problem of {\em embedding} an image into the latent space of Denoising Diffusion Models, that is finding a suitable ``noisy'' image whose denoising results in the original image. We particularly focus on Denoising Diffusion Implicit Models due to the deterministic nature of their reverse diffusion process. As a side result of our investigation, we gain a deeper insight into the structure of the latent space of diffusion models, opening interesting perspectives on its exploration, the definition of semantic trajectories, and the manipulation/conditioning of encodings for editing purposes. A particularly interesting property highlighted by our research, which is also characteristic of this class of generative models, is the independence of the latent representation from the networks implementing the reverse diffusion process. In other words, a common seed passed to different networks (each trained on the same dataset), eventually results in identical images.}


\maketitle

\section{Introduction}\label{sec:introduction}
Denoising Diffusion Models (DDM) \cite{DDPM} are rapidly imposing as the new state-of-the-art technology in the field of deep generative modeling, challenging the role held so far by Generative Adversarial Networks \cite{Diff_vs_GAN}. The impressive text-to-image generation capability shown by models like DALL$\cdot$E \!2 \cite{DALLE2} and Imagen \cite{Imagen}, recently extended to videos in \cite{VDM}, clearly proved the qualities of this technique, comprising excellent image synthesis quality, good sampling diversity, high sensibility and easiness of conditioning, stability of training and good scalability.

In very rough terms, a diffusion model trains a single network to denoise images
with a parametric amount of noise, and generates images by iteratively denoising 
pure random noise. This latter process is traditionally called {\em reverse diffusion} since it is meant to ``invert'' the {\em direct diffusion} process consisting in adding noise. In the important case of Implicit Diffusion models \cite{DDIM}, reverse diffusion is deterministic, but obviously not injective: many noisy images can be denoised to a single common result. Let us call $emb(x)$ (embedding of $x$) the set of points whose reverse diffusion generate $x$. The problems we are interested in are investigating the shape of $emb(x)$ (e.g. is it a connected, convex space?), finding a ``canonical'' element in it (i.e. a sort of center of gravity) and, in case such a canonical element exists, finding an efficient way to compute it. This would allow us to embed an arbitrary image into the ``latent'' space of a diffusion model, providing functionality similar to GAN-recoders (see Section~\ref{sec:related}), or to encoders in the case of Variational AutoEncoders (\cite{VAEKingma},\cite{VAEGreen}).

\begin{figure}[ht]
\begin{center}
\includegraphics[width=\columnwidth]{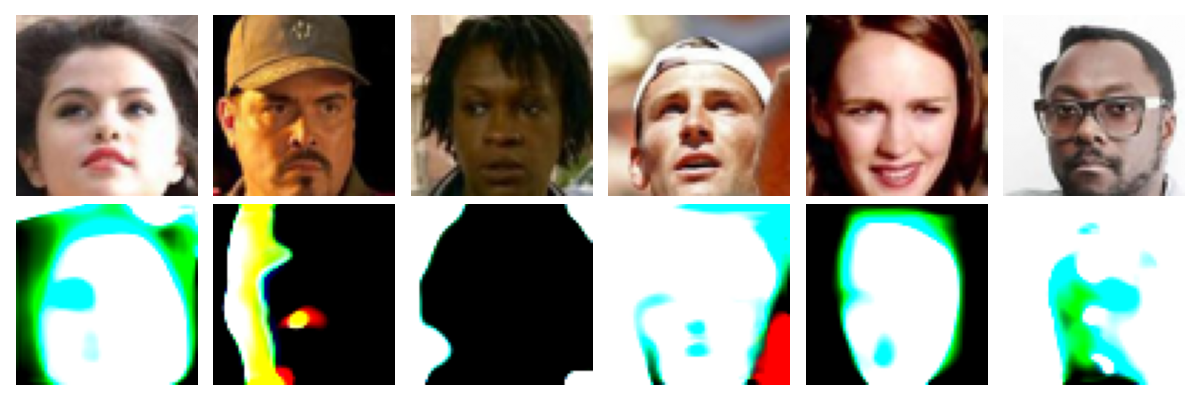}
\caption{Examples of faces obtained using the image itself as input to the reverse diffusion process. The input signal is clearly far too strong.
\label{fig:identity_generation}}
\end{center}
\end{figure}

Since reverse diffusion is the ``inversion'' of the diffusion process, it might be natural to expect $emb(x)$ to be composed by noisy versions of $x$, and that the canonical element we are looking for could be $x$ itself. This is not the case: indeed, $x$ does not seems to belong to $emb(x)$. Figure~\ref{fig:identity_generation} details some examples of the output obtained by using the image itself as input to the reverse diffusion process.

\hspace{-.4cm}\begin{minipage}{.45\textwidth}
Since the input signal is clearly too strong, we may be tempted to reduce it 
using a multiplicative factor equal to the minimum signal rate used to train the 
denoising network ($0.02$ in our case), or a similarly low value. Examples of results are shown in Figure~\ref{fig:weak_identity_generation}. Although some macroscopic aspects of the original image like orientation and illumination are roughly preserved, most of the information is not embedded in these seeds: scaling does not result in a reasonable embedding. We also attempted to inject some additional noise into the initial seed, hoping to obtain a more entropic signal that is similar to the typical input of the reverse diffusion process, but this merely resulted in a less deterministic output.

Therefore, the embedding problem is both far from trivial and very interesting. Understanding the embedding would give us a better grasp of the reverse diffusion progress, as well as a deeper, semantic insight into the structure of its latent space.
\end{minipage}\hspace{.3cm}
\begin{minipage}{.5\textwidth}
\begin{figure}[H]
\begin{center}
\includegraphics[width=\textwidth]{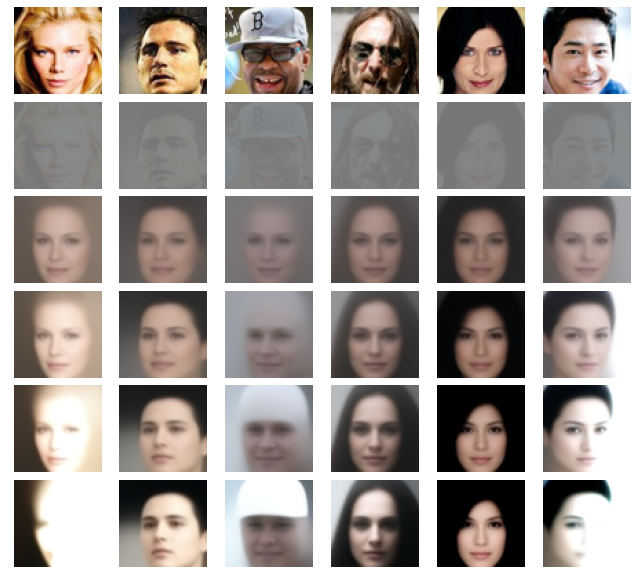}
\caption{Examples of faces obtained using a weak version of the image itself as
input signal. The first row shows the original image, while the second shows its
weak version, which has been scaled by the minimum signal rate used to train the denoising network (0.02). This weaker image constitutes the initial seed. In the following four rows, we see the reconstructions obtained through reverse diffusion
from the initial seed and progressively stronger versions of it, varying the signal rate between $0.02$ and $0.08$.
\label{fig:weak_identity_generation}}
\end{center}
\end{figure}
\end{minipage}
\smallskip

\begin{figure}[ht]
\begin{center}
\includegraphics[width=\columnwidth]{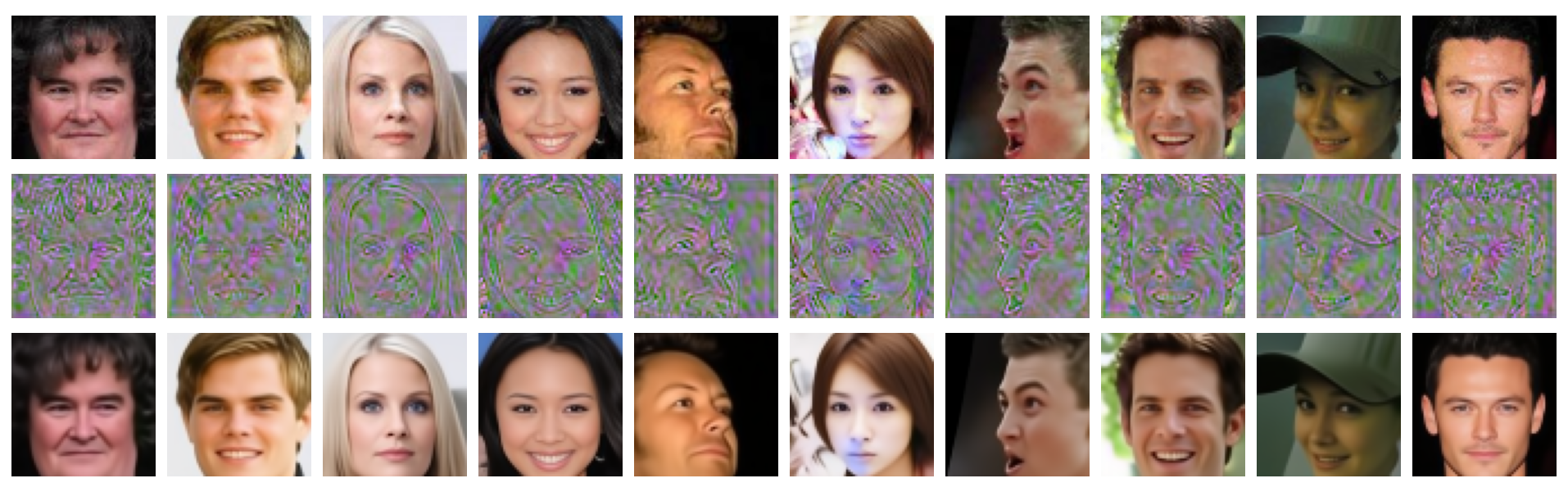}
\caption{CelebA: examples of face embeddings (second row) and reconstructions (third row). No cherry-picking.}
\label{fig:embedding_final}
\end{center}
\end{figure}

Our approaches to the embedding problem are discussed in Section~\ref{sec:embedding}. Overall, we find that we can obtain pretty good results by directly training a Neural Network to compute a kind of ``canonical" seed (see Figure~\ref{fig:embedding_final}).

The reconstruction quality is very high, with an MSE of around 0.0015 in the case of CelebA \cite{celeba}. More detailed values are provided in Section~\ref{sec:recoder}. 

\begin{figure}[ht]
\begin{center}
\begin{tabular}{cccccc}
\includegraphics[width=.166\columnwidth]{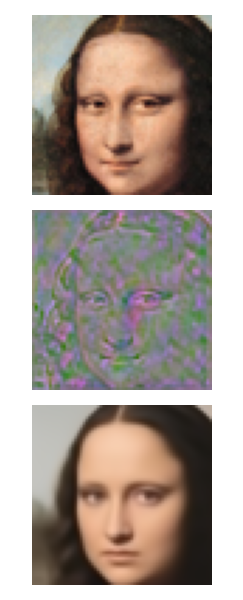} & \hspace{-.5cm}\includegraphics[width=.166\columnwidth]{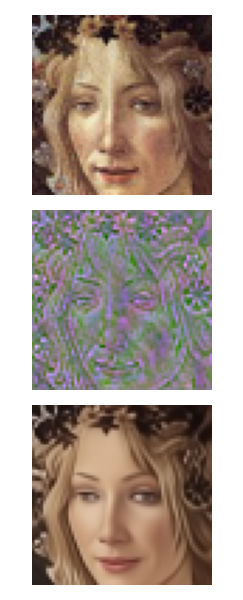} & 
\hspace{-.5cm}\includegraphics[width=.166\columnwidth]{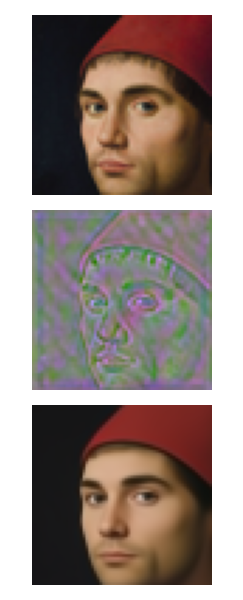} & 
\hspace{-.5cm}\includegraphics[width=.166\columnwidth]{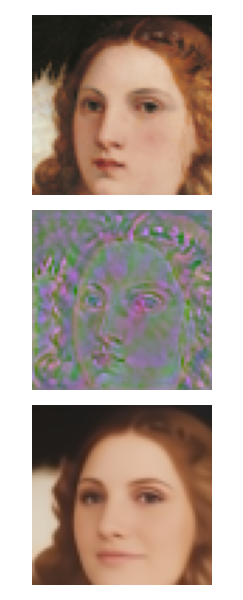} & 
\hspace{-.5cm}\includegraphics[width=.166\columnwidth]{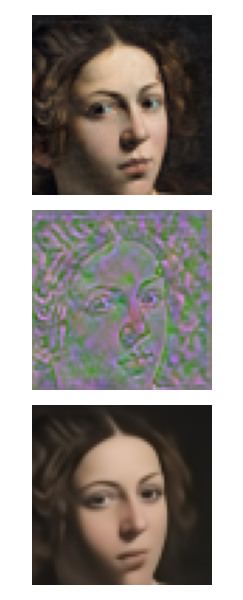} & 
\hspace{-.5cm}\includegraphics[width=.166\columnwidth]{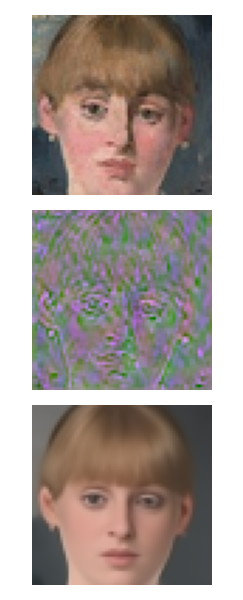}\\
Leonardo & \hspace{-.5cm}Botticelli & \hspace{-.5cm}Antonello & \hspace{-.5cm}Tiziano &\hspace{-.5cm}Caravaggio & \hspace{-.5cm}Manet\\
& & \hspace{-.5cm}da Messina & & &

\end{tabular}
\caption{Reification of portraits. The portrait is first embedded into the latent space, and then pulled back into the data manifold.}
\label{fig:reification}
\end{center}
\end{figure}

A typical application of the embedding process consists in transforming a signal into an element of the data manifold sufficiently close to it (the same principle behind denoising autoencoders). An amusing utilization is for the reification of artistic portraits, as exemplified in Figure~\ref{fig:reification}.

\begin{figure}[H]
\begin{center}
\begin{tabular}{cccc}
\hspace{-.7cm}\includegraphics[width=.2\columnwidth]{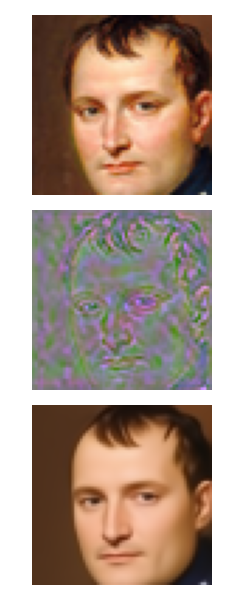} & 
\hspace{-.5cm}\includegraphics[width=.2\columnwidth]{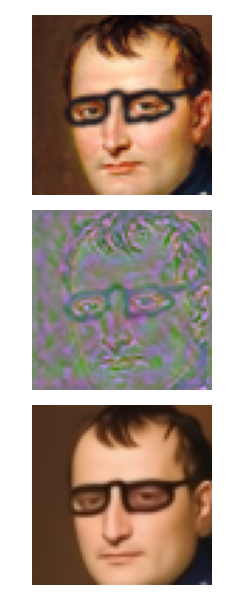} & 
\hspace{-.7cm}\includegraphics[width=.2\columnwidth]{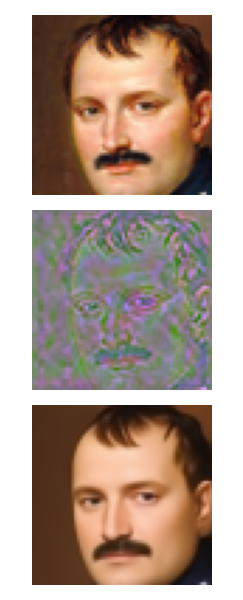} & 
\hspace{-.7cm}\includegraphics[width=.2\columnwidth]{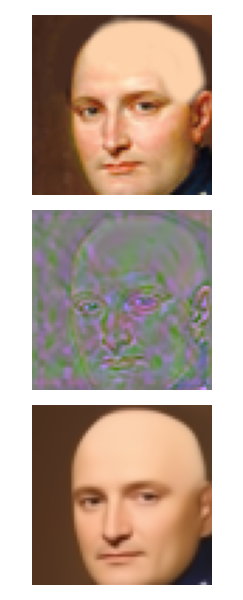}
\end{tabular}
\hspace*{-1.5cm}\caption{Scribbles over David's Napoleon.}
\label{fig:scribbles}
\end{center}
\end{figure}

\vspace{-.3cm}
Another interesting possibility is that of making sketchy modifications to 
an image (a portrait, or a photograph) and delegating to the embedder-generator pair the burden of integrating them in the original picture in a satisfactory way (see Figure~\ref{fig:scribbles}).

\subsection{Structure of the Work}
The article is structured as follows. In Section~\ref{sec:related}, we discuss related works, mostly focusing on the embedding problem for Generative Adversarial Networks. Section~\ref{sec:theory} is devoted to formally introducing the notion of Denoising Diffusion Models, in addition to the deterministic variant of Denoising Diffusion Implicit Models we are particularly interested in. In the same section, we also discuss an intuitive interpretation of denoising diffusion models in terms of a ``gravitational analogy'' (Section~\ref{sec:gravitational}), which drove many of our investigations and plays an important role in understanding the structure of datapoint embeddings. A major consequence of this interpretation, which to the best of our knowledge has never been pointed out before, is the {\em invariance of the latent space with respect to different models}: a given seed, passed as input to different models, always produces the same image. 
In Section~\ref{sec:denoising_architecture}, we provide architectural details about our implementation of the Denoising Diffusion model. Our methodology to address the embedding problem is discussed in Section~\ref{sec:embedding}.
Two main approaches have been considered, one based on a gradient descent technique, which allows us to synthesize large clouds of different seeds in the embedding space of specific data points (Section~\ref{sec:recoder}), and another one based on training a neural network to compute a single ``canonical'' seed for the given image: essentially, a sort of encoder.
Conclusions and future works are discussed in Section~\ref{sec:conclusions}.
\smallskip

\textbf{Code}. The source code of the experiments described in this work is freely available at the GitHub repository \href{https://github.com/asperti/Embedding-in-Diffusion-Models}{https://github.com/asperti/Embedding-in-Diffusion-Models}, along with links to weights for pre-trained models.

\section{Related Works}
\label{sec:related}
The embedding problem has been extensively investigated in the case of 
Generative Adversarial Networks (GANs) \cite{gan_inversion_survey}. Similarly 
to Denoising Generative Models, GANs lack a direct encoding process of the original input sample into the latent space.

Several approaches to inversion have been investigated \cite{perarnau2016invertible, bau2020semantic, daras2020your, anirudh2020mimicgan}, mostly with the purpose of editing. The most common approaches are based on synthesis of the latent encoding via gradient descent techniques \cite{gan_inversion_optimization}, or by training a suitable neural network to produce encodings able to reconstruct the original input with sufficient approximation. While the former technique generally tends to achieve better reconstruction errors, the latter has faster inference times and can take advantage of the fact that, since a GAN produces an infinite stream of training data, over-fitting is much less likely. Hybrid methods combining 
both techniques have also been explored \cite{gan_inversion_first_hybrid, gan_inversion_hybrid_2}.

Recent works have mostly focused on the inversion of the popular StyleGAN and its successors \cite{karras2019style, karras2020analyzing, karras2021alias}, building on previous work with a variety of inversion structures and minimization objectives, or aiming to generalize/transfer to arbitrary datasets \cite{stylegan_inversion, collins2020editing, abdal2020image2stylegan, stylegan_overparameterization, alaluf2022hyperstyle}.

As we already mentioned, the typical application of the embedding is for exploration of the latent space, either for disentanglement purposes or in view of editing; the two issues are in fact tightly intertwined, since knowledge about semantically meaningful directions (e.g. color, pose, shape) can be exploited to tweak an image with the desired features. For instance, InterFaceGAN \cite{Shen22} uses regression techniques to find a hyperplane in the latent space whose normal vector allows for a gradual modification of the feature. Further work based on this idea searches for these directions as an iterative or an optimization problem \cite{li2021interpreting} and also extends it to controllable walks in the latent space \cite{li2021discovering}. In the same vein, \cite{Diffusion_latent_space} studies the feature space of the U-Net bottleneck of the diffusion model, finding that it can be used as an alternative latent space with highly semantic directions.

Another important application of embeddings is for the comparison of the latent space of different generative models \cite{comparingNCAA}: having the possibility to embed the same image in different spaces allows us to create a supervised dataset suitable to learn direct mappings from one space to another.

In the realm of diffusion models, much work has been done on the refinement of the reverse diffusion process \cite{nichol2021improved, conditioning_denoising, Diff_vs_GAN}, but relatively little attention has been so far devoted to its inversion. DALL$\cdot$E \!2 \cite{DALLE2} relies on a form of image embedding, but this is a pre-trained contrastive model not learnt as the inversion of the generative task. An additional difference with respect to our work is that we are also interested in investigating and understanding the {\em structure} of the embedding cloud for each image since it could highlight the organization of the latent space and the sampling process. 


Finally, in the context of text-conditioned generative models, interesting attempts to invert not just the image but a user-provided concept have been 
investigated in \cite{textualInversion}. The concept is represented as a new pseudo-word in the model's vocabulary, which can be then used as part of a prompt (e.g. ``a flower in the style of $S_*$'', where $S_*$ refers to an image). The mapping is achieved by optimizing the conditioning vector in order to minimize the reconstruction error (similarly to the technique described in Section~\ref{sec:gradient_descent}). A similar approach is used in \cite{dreamArtist}. 




\section{Denoising Diffusion Models}
\label{sec:theory}

In this section, we provide a general overview of diffusion models from a mathematical perspective.

Consider a distribution $q(x_0)$ generating the data. Generative models aim to find a parameter vector $\theta$ such that the distribution $p_\theta(x_0)$, parameterized by a neural network, approximates $q(x_0)$ as accurately as possible. In Denoising Diffusion Probabilistic Models (DDPM) \cite{DDPM}, the generative distribution $p_\theta(x_0)$ is assumed to have the form
\begin{equation}
    p_\theta(x_0) = \int p_\theta(x_{0:T}) dx_{1:T}
\end{equation}
for a given time range horizon $T > 0$, where 
\begin{equation}
    p_\theta(x_{0:T}) = p_\theta(x_T) \prod_{t=1}^T p_\theta(x_{t-1}\vert x_t)
\end{equation}
with $p_\theta(x_T) = \mathcal{N}(x_T \vert 0; I)$ and $p_\theta(x_{t-1}\vert x_t) = \mathcal{N}(x_{t-1} \vert \mu_\theta(x_t, \alpha_t); \sigma_t^2 I)$. Similarly, the diffusion model $q(x_{0:T})$ is considered to be a Markov chain of the form
\begin{equation}
    q(x_t \vert x_{t-1}) = \mathcal{N}\Biggl(x_t \Bigg\vert \sqrt{\frac{\alpha_t}{\alpha_{t-1}}} x_{t-1}; \Bigl(1 - \frac{\alpha_t}{\alpha_{t-1}}\Bigl) \cdot I\Biggr)
\end{equation}
with $\{ \alpha_t \}_{t \in [0, T]}$ being a decreasing sequence in the interval $[0, 1]$. The parameters of the generative model $p_\theta(x_0)$ are then trained to fit $q(x_0)$ by minimizing the negative Evidence Lower BOund (ELBO) loss, defined as
\begin{equation}
    \mathcal{L}(\theta) = - \E_{q(x_{0:T})} [ \log p_\theta(x_{0:T}) - \log q(x_{1:T}) ].
\end{equation}
The ELBO loss can be rewritten in a computable form by noticing that, as a consequence of Bayes' Theorem,  $q(x_{t-1} \vert x_t, x_0) = \mathcal{N}(x_{t-1} \vert \tilde{\mu}(x_t, x_0); \sigma_q^2 \cdot I)$. Consequently,
\begin{equation}\label{eq:ELBO_formulation}
    \mathcal{L}(\theta) = \sum_{t=1}^T \gamma_t \E_{q(x_t \vert x_0)} \Bigl[ \| \mu_\theta(x_t, \alpha_t) - \tilde{\mu}(x_t, x_0) \|_2^2 \Bigr]
\end{equation}
which can be interpreted as the weighted mean squared error between the reconstructed image from $p_\theta(x_t \vert x_0)$ and the true image obtained by the reverse diffusion process $q(x_{t-1} \vert x_t, x_0)$ for each time $t$. \\

In \cite{DDIM}, the authors considered a non-Markovian reverse diffusion process (also called inference distribution)
\begin{equation}
    q_\sigma (x_{1:T} \vert x_0) = q_\sigma(x_T \vert x_0) \prod_{t=2}^T q_\sigma (x_{t-1} \vert x_t, x_0)
\end{equation}
where $q_\sigma(x_T \vert x_0) = \mathcal{N}(x_T \vert \sqrt{\alpha_T} x_0, (1 - \alpha_T) \cdot I)$, and

\begin{align}\label{eq:non_markovian_reverse_diffusion}
    q_\sigma (x_{t-1} \vert x_t, x_0) = \mathcal{N} \Bigl( x_{t-1} \Big \vert \mu_{\sigma_t}(x_0, \alpha_{t-1}); \sigma_t^2 \cdot I \Bigr)
\end{align}
with 
\begin{align}
    \mu_{\sigma_t}(x_0, \alpha_{t-1}) = \sqrt{\alpha_{t-1}} x_0 + \sqrt{1 - \alpha_{t-1} - \sigma_t^2} \cdot \frac{x_t - \sqrt{\alpha_t} x_0}{\sqrt{1 - \alpha_t}}.
\end{align}
This construction implies that the forward process is no longer Markovian, since it depends both on the starting point $x_0$ and on $x_{t-1}$. Moreover, \cite{DDIM} proved that, with this choice of $q_\sigma(x_{1:T} \vert x_0)$, the marginal distribution $q_\sigma(x_t\vert x_0) = \mathcal{N}(x_t \vert \sqrt{\alpha_t} x_0; (1 - \alpha_t) \cdot I)$, recovers the same marginals as in DDPM, which implies that $x_t$ can be diffused from $x_0$ and $\alpha_t$ by generating a realization of normally distributed noise $\epsilon_t \sim \mathcal{N}(\epsilon_t \vert 0; I)$ and defining
\begin{align}\label{eq:sampling_xt_from_x0}
    x_t = \sqrt{\alpha_t} x_0 + \sqrt{1 - \alpha_t} \epsilon_t.
\end{align}
Note that when in Equation \eqref{eq:non_markovian_reverse_diffusion} $\sigma_t = 0$, the reverse diffusion $q_\sigma(x_{t-1} \vert x_t, x_0)$ becomes deterministic. With such a choice of $\sigma_t$, the resulting model is named Denoising Diffusion Implicit Models (DDIM) by the authors in \cite{DDIM}. Interestingly, in DDIM, the parameters of the generative model $p_\theta(x_{t-1} \vert x_t)$ can be simply optimized by training a neural network $\epsilon_\theta^{(t)}(x_t, \alpha_t)$ to map a given $x_t$ to an estimate of the noise $\epsilon_t$ added to $x_0$ to construct $x_t$ as in \eqref{eq:sampling_xt_from_x0}. Consequently, $p_\theta(x_{t-1} \vert x_t)$ becomes a $\delta_{f_\theta^{(t)}}$, where
\begin{align}\label{eq:nn_DDIM}
    f_\theta^{(t)}(x_t, \alpha_t) = \frac{x_t - \sqrt{1 - \alpha_t} \epsilon_\theta^{(t)}(x_t, \alpha_t)}{\sqrt{\alpha_t}}.
\end{align}
Intuitively, the network in \eqref{eq:nn_DDIM} is just a denoiser that takes as input the noisy image $x_t$ and the variance of the noise $\alpha_t$ and returns an estimate of the denoised solution $x_0$. In DDIM, one can generate new data by first considering random Gaussian noise $x_T \sim p_\theta(x_T)$ with $\alpha_T = 1$. Then, $x_T$ is processed by $f_\theta^{(T)}(x_T, \alpha_T)$ to generate an estimation of $x_0$, which is then corrupted again by the reverse diffusion $q(x_{T-1} \vert x_T, f_\theta^{(T)}(x_T, \alpha_T))$. This process is repeated until a new datum $x_0$ is generated by $f_\theta^{(1)}(x_1, \alpha_1)$. \\

The sampling procedure of DDIM generates a trajectory $\{ x_T, x_{T-1}, \dots, x_0 \}$ in the image space. 
In \cite{song2020score,khrulkov2022understanding} the authors found that the (stochastic) mapping from $x_T$ to $x_0$ in DDPM follows a score-based stochastic differential equation (SDE), where the dynamic is governed by terms related to the gradient of the ground-truth probability distribution from which the true data is generated. The sampling procedure for DDIM can be
obtained by discretizing the deterministic \emph{probability flow} \cite{song2020score} associated with this dynamics.
Consequently, training a DDIM model leads to an approximation of the score function of the ground-truth distribution. 

\subsection{The Diffusion Schedule}
An important aspect in implementing diffusion models is the choice of the diffusion noise $\{ \alpha_t \}_{t=1}^T$, defining the mean and the variance of $q(x_t \vert x_0)$. In \cite{DDPM}, the authors showed that the diffusion process $q(x_t \vert x_0)$ converges to a normal distribution if and only if $\alpha_T \approx 0$. Moreover, to improve the generation quality, $\alpha_t$ has to be chosen such that it slowly decays to 0. 
%
%

%
%
%

\begin{center}
\begin{figure}[hb]
    \includegraphics[width=\linewidth]{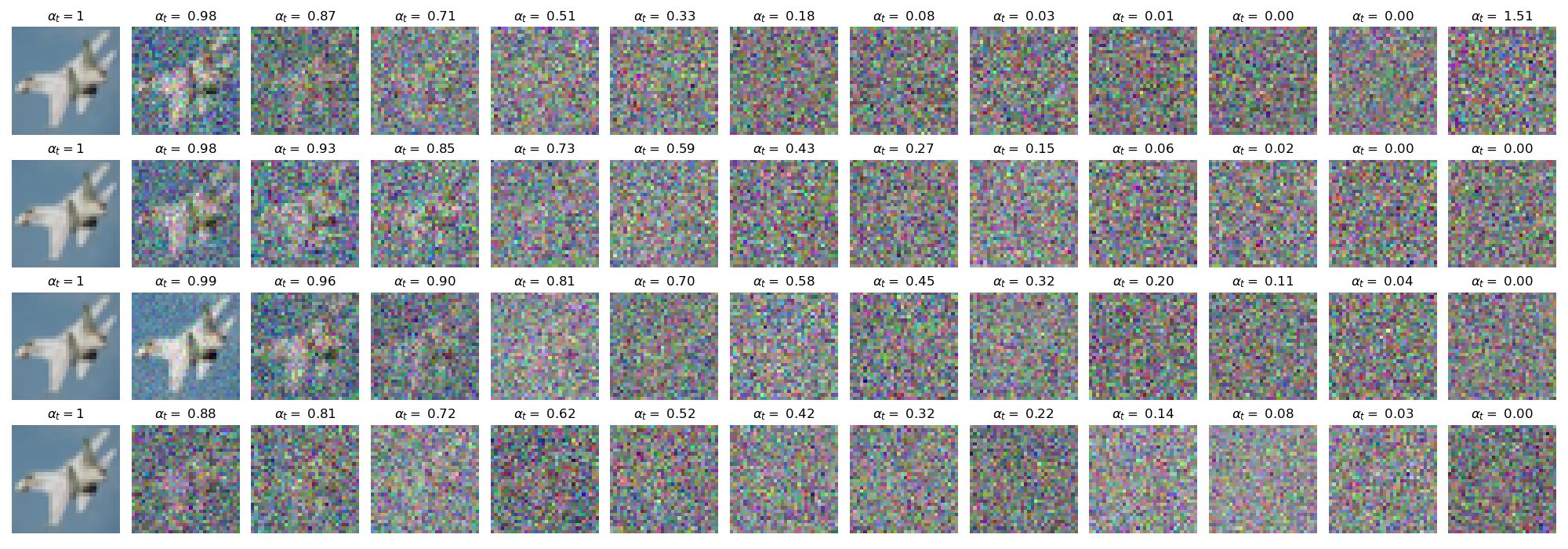}
    \caption{Increasing noise added by the different scheduling: in order, from top to bottom, linear, quadratic, cosine, and continuous cosine schedules.  For each row, from left to right, the time $t$ is increased linearly from $0$ to $T$. The corresponding $\alpha_t$ for each schedule and for any $t$ is reported above each image.}
    \label{fig:schedule_image}
\end{figure}
\end{center}

\hspace{-.5cm}\begin{minipage}{5.5cm}
The specific choice for the sequence $\alpha_t$ defines the so-called \emph{diffusion schedule}. 

In \cite{DDPM}, the authors proposed to use linear or quadratic schedules.
This choice was criticized in \cite{kingma2021variational,nichol2021improved}
since it exhibits a too steep decrease during the first time steps, causing difficulties during generation for the neural network model. 
To remedy this situation, alternative scheduling functions with a gentler decrease have been proposed in the literature, such as the {\em cosine} or {\em continuous cosine} schedule.
The behavior of all these functions is compared in 
Figure \ref{fig:schedule_plot}.

\end{minipage}\hspace{.5cm}
\begin{minipage}{5.6cm}
\begin{figure}[H]
\begin{center}
\includegraphics[width=\columnwidth]{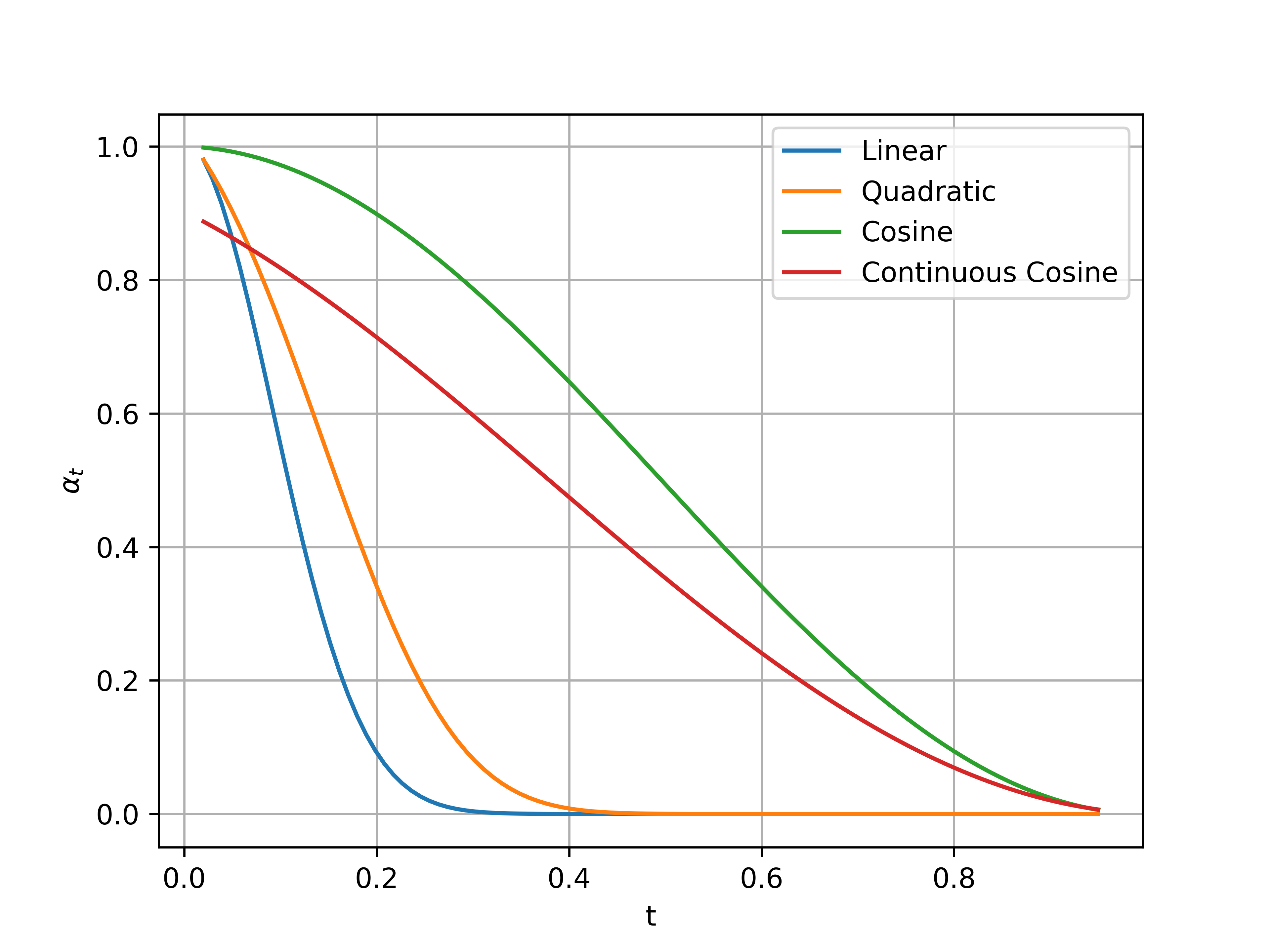}
\caption{Comparison of different schedules. Generation is better if noise variance does not change too abruptly, so cosine and continuous cosine schedules usually work better than the linear or quadratic ones.}
\label{fig:schedule_plot}
\end{center}
\end{figure}
\end{minipage}
The quantity of noise added by each schedule is also represented in Figure \ref{fig:schedule_image}, where a single image is injected with increasing noise according to the given schedule. It is not hard to see that the cosine and the continuous cosine schedules exhibit a more uniform transaction between the original image and the pure noise.

\subsection{The Gravitational Analogy}
\label{sec:gravitational}
Similarly to other generative models, developing an intuition of the actual behavior of diffusion models (and of the mapping from a latent encoding to its visible outcome) can be challenging.
In this section, we propose a simple gravitational analogy that we found extremely 
useful to get an intuitive grasp of these models, and which suggested us some 
interesting conjectures about the actual shape of the embedding clouds for each object.


Simply stated, the idea is the following. You should think of the datapoints as corps with a gravitational attraction. Regions of the space where the data manifold has high probability are equivalent to regions with high density. The denoising model essentially learns the gravitational map induced over the full space: any single point of the space gets mapped to the point where it would naturally ``land" if subject to the ``attraction" of the data manifold. 

In more explicit terms, {\em any} point $z$ of the space can be seen as a noisy version of {\em any} point $x$ in the dataset. The ``attraction" exerted by $x$ on $z$ (i.e. the loss) is directly proportional to their distance, usually an absolute or quadratic error.

\hspace{-.5cm}\begin{minipage}{5cm}
However, the probability to train the network to reconstruct $x$ from $z$ has a
Gaussian distribution $\mathcal{N}(z \vert x ; \sigma_z \cdot I)$, with $\sigma_z$ depending on the denoising step.
Hence, the {\em weighted attraction} exerted by $x$ on $z$ at each step is
\begin{equation}
\mathcal{N}(z \vert x ; \sigma_z \cdot I) \cdot \|x-z\|
\label{eq:gravitation_force}
\end{equation}
To get a grasp of the phenomenon, in Figure~\ref{fig:gravitational_analogy} we compare the gravitational low for a corp $x$ with the {\em weighted attraction} reported in Equation \eqref{eq:gravitation_force}, under the assumption that the variance $\sigma$ has to be compared with the radius of the corp (with constant density, for simplicity).
\end{minipage}\hspace{1cm}
\begin{minipage}{5.5cm}
\begin{figure}[H]
\begin{center}
\includegraphics[width=\columnwidth]{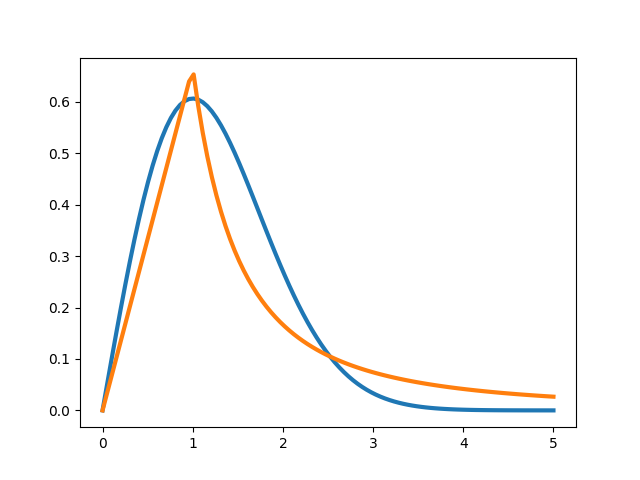}
\caption{Gravitational analogy. The orange line is the usual gravitational low for a corp with radius 1 and constant density. The blue line is a weighted loss $\mathcal{N}(z \vert x; 1)\cdot \|x-z\|_1$. The two lines have been rescaled to have an equal integral. 
}
\label{fig:gravitational_analogy}
\end{center}
\end{figure}

\end{minipage}

\begin{figure}[h]
\begin{center}
\begin{tabular}{cccc}
\hspace{-.5cm}\includegraphics[width=.24\columnwidth]{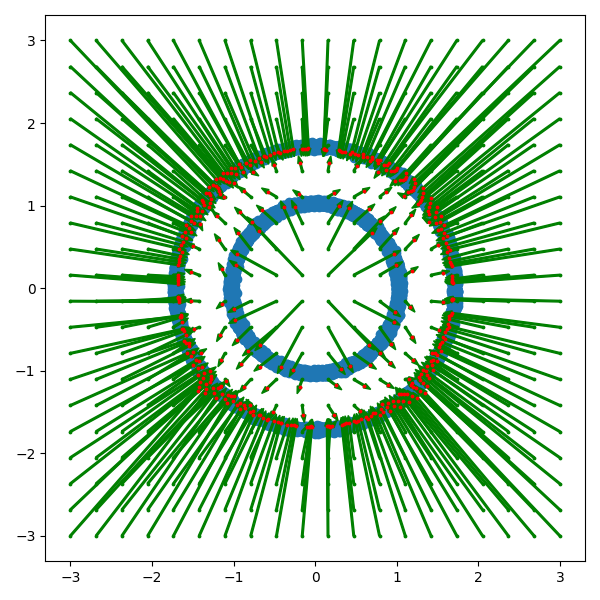} &
\hspace{-.5cm}\includegraphics[width=.24\columnwidth]{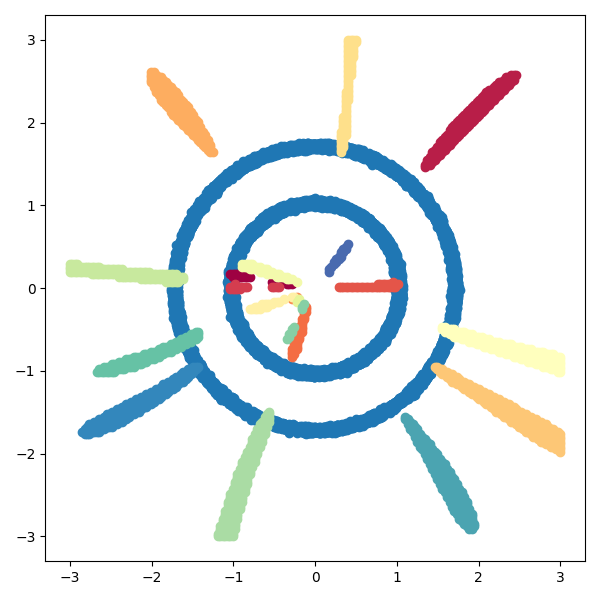} &
\hspace{-.5cm}\includegraphics[width=.24\columnwidth]{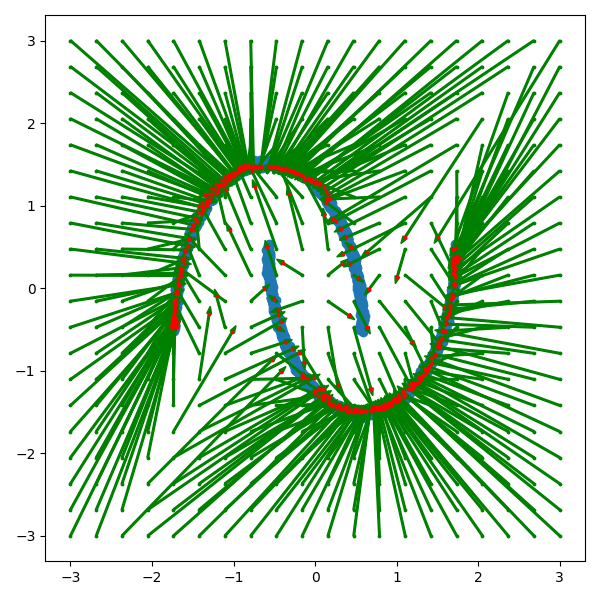} &
\hspace{-.5cm}\includegraphics[width=.24\columnwidth]{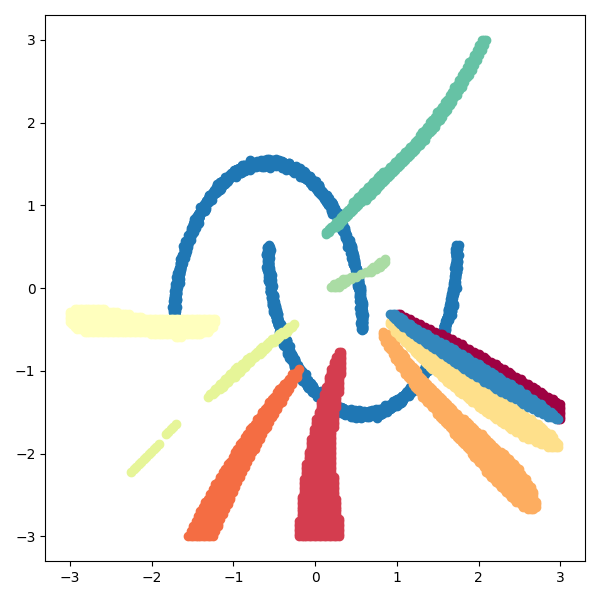}\\
(a) circles & (b) two moons & (c) circle & (d) two moons
\end{tabular}
\caption{Gravitational map and embeddings for the ``circles" (a,b) and ``two moons" (c,d) datasets. Datapoints are in blue. We consider a dense grid of seeds in the latent space, depicted in green. To visualize the maps (a) and (c) we draw
an arrow pointing from each seed to the corresponding point generated by reverse diffusion (in red). To visualize embeddings (b) and (d) we consider a set of elements in the datasets, and for each element $x$ we consider all points in the grid generating a sample $\hat{x}$ sufficiently close to $x$. 
}
\label{fig:gravitational_maps}
\end{center}
\end{figure}

According to the gravitational analogy, the embedding space $emb(x)$ of each datapoint $x$ should essentially coincide with the set of points in the space corresponding to trajectories ending in $x$. We can study this hypothesis on synthetic datasets. In Figure~\ref{fig:gravitational_maps} we show the gravitational map for the well-known ``circle" (a) and ``two moons" datasets (b); examples of embeddings are given in figures (c) and (d). 

From the pictures, it is clear in which way the model ``fills the space", that is associating to each datapoint $x$ all ``trajectories" landing in $x$. The trajectories are almost straight and oriented along directions orthogonal to the data manifold. We believe that this behavior can be formally understood by exploiting the dynamics of the trajectories introduced in \cite{song2020score}, as mentioned in Section \ref{sec:theory}. We aim to deeply investigate those aspects in a future work.

The most striking consequence of the ``gravitational" interpretation is, however, the independence of the latent encoding from the neural network or its training: the gravitational map only depends on the data manifold and it is unique, so distinct networks or different trainings of the same network, if successful, should eventually end up with the same results. This seems miraculous: if we pick a random seed in an almost immense space, and pass it as input to two diffusion (deterministic) models for the same dataset, they should generate essentially \emph{identical} images.

\begin{figure}[h]
\begin{center}
\begin{subfigure}[h]{\textwidth}
  \centering
  \includegraphics[width=0.6\linewidth]{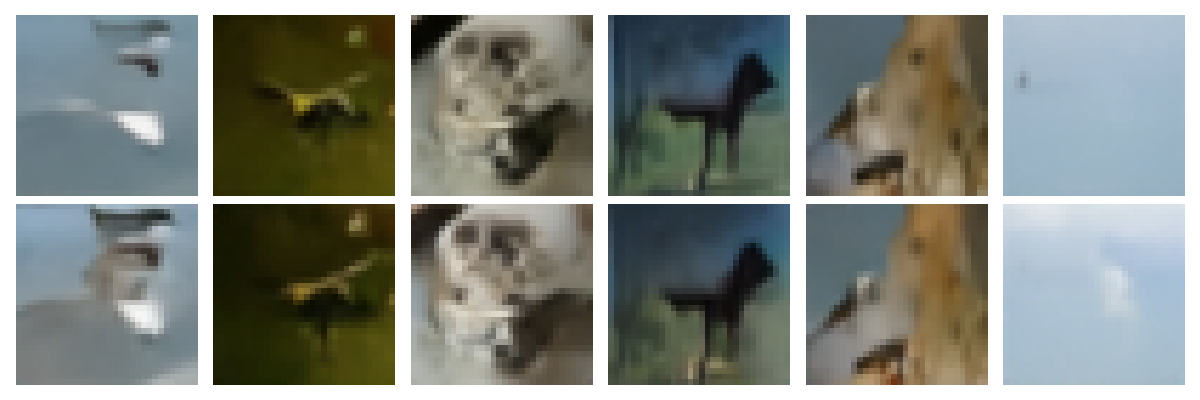}  
\end{subfigure}
\begin{subfigure}[h]{\textwidth}
  \centering
  \includegraphics[width=0.6\linewidth]{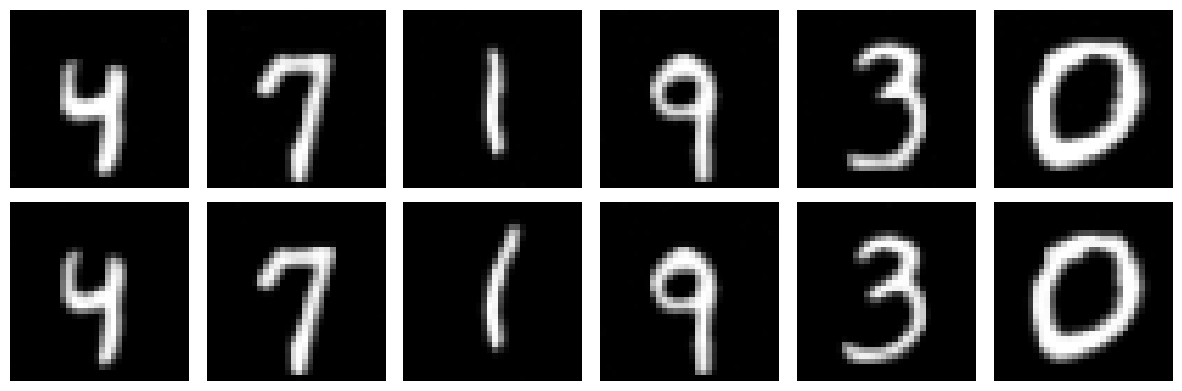}  
\end{subfigure}
\begin{subfigure}[h]{\textwidth}
  \centering
  \includegraphics[width=0.6\linewidth]{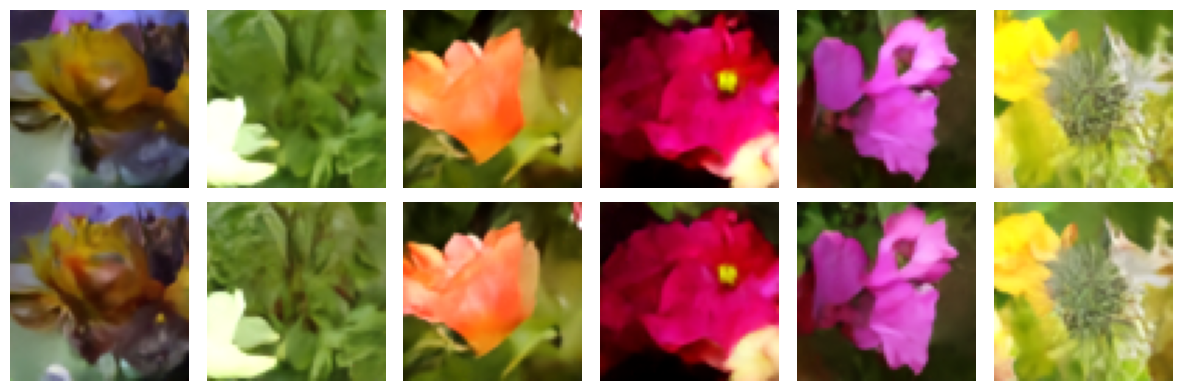}  
\end{subfigure}
\caption{Uniqueness of the generative model. Different diffusion models generate essentially identical images when fed with the same seed. The two models have dissimilar architectures, with a higher number of parameters for the first row and lower for the second. The training sets are CIFAR10 (top), MNIST (middle), and Oxford Flowers 102 (bottom); seeds have been randomly generated.}
\label{fig:uniqueness}
\end{center}
\end{figure}

This can be easily experimentally verified. In Figure~\ref{fig:uniqueness} we show images generated from two different models trained over CIFAR10 starting from the same set of seeds: they are practically identical.

The fact that the same encoding works for different models seems to be peculiar to 
this kind of generative models. In \cite{comparingNCAA}, it was observed that we can essentially pass from a latent space to another of different generative models
with a simple {\em linear map}: however, an identity or even a permutation of latent variables does not usually suffice\footnote{It remains to be checked if imposing a spatial structure to the latent space of GANs and VAEs is enough to induce uniqueness in that case too. We plan to investigate this issue in a forthcoming work.}. 

While the uniqueness of the latent space is, in our opinion, a major discovery, it is not the main focus of this article, and we plan to conduct a more exhaustive and principled investigation of this property in future works.

\section{Denoising Architecture}
\label{sec:denoising_architecture}

The pseudocodes explaining training and sampling for diffusion models are respectively given in Algorithms~\ref{algorithm1} and \ref{algorithm2} below.

\noindent
\begin{minipage}{0.48\textwidth}
\begin{algorithm}[H]
    \centering
    \caption{Training}\label{algorithm1}
    \begin{algorithmic}[1]
        \Repeat
        \State $x_0 \sim q(x_0) $
        \State $t \sim $Uniform({1,..,T})
        \State $\epsilon \sim \mathcal{N}(0;I) $
        \State Take gradient descent step on 
        $\lvert \lvert \epsilon - \epsilon_{\theta} (\sqrt{\alpha_t} x_0 + \sqrt{1\!-\! \alpha_t} \epsilon, \alpha_t )\rvert \rvert^2 $
        \Until converged
    \end{algorithmic}
\end{algorithm}
\end{minipage}
\hfill
\begin{minipage}{0.50\textwidth}\vspace{-.4cm}
\begin{algorithm}[H]
    \centering
    \caption{Sampling}\label{algorithm2}
    \begin{algorithmic}[1]
        \State $x_T \sim \mathcal{N}(0,I) $
        \State \textbf{for} {$t = T,...,1$} \textbf{do}
        \State $t \sim $Uniform({1,..,T})
        \State $z \sim \mathcal{N}(0;I)\; \mathit{if}\; t>1, \mathit{else}$ \textbf{z}=\textbf{0} 
        \State  
        $x_{t-1}\! =\! \frac{1}{\sqrt{\alpha_t}} (x_t - \frac{1-\alpha_t}{\sqrt{1 - \alpha_t}} \epsilon_\theta(x_t,\alpha_t))\! +\! z$
        \State \textbf{end for}
        
    \end{algorithmic}
\end{algorithm}
\end{minipage}
\bigskip

As explained in Section~\ref{sec:theory}, they are iterative algorithms; the only trainable component is the denoising network $\epsilon_\theta(x_t, \alpha_t)$, which takes as input a noisy image $x_t$ and a noise variance $\alpha_t$, and tries to estimate the noise present in the image. This model is trained as a traditional denoising network, taking a sample $x_0$ from the dataset, corrupting it with the expected amount of random noise, and trying to identify the noise in the noisy image. 

As a denoising network, it is quite standard to consider a conditional variant of the U-Net. This is a very popular network architecture originally proposed for semantic segmentation \cite{U-net} and subsequently applied to a variety of image manipulation tasks. In general, the network is structured with a downsample sequence of layers followed by an upsample sequence, with skip connections added between the layers of the same size. 

To improve the sensibility of the network to the noise variance, $\alpha_t$ is taken as input, which is then embedded using an ad-hoc sinusoidal transformation by splitting the value in a set of frequencies, in a way similar to positional encodings in Transformers \cite{attention}.
The embedded noise variance is then vectorized and concatenated to the noisy images along the channel axes before being passed to the U-Net. This can be done for
each convolution blocks separately, or just at the starting layer; we adopted the latter solution due to its simplicity and the fact that it does not seem to entail any loss in performance.

Having worked with a variety of datasets, we used slightly different implementations of the previously described model. The U-Net is usually parameterized by specifying the number of downsampling blocks, and the number of channels for each block;
the upsampling structure is symmetric. The spatial dimension does not need to be specified, since it is inferred from the input. Therefore, the whole structure of a U-Net is essentially encoded in a single list such as [32, 64, 96, 128] jointly
expressing the number of downsampling blocks (4, in this case), and the respective number of channels (usually increasing as we decrease the spatial dimension).

For our experiments, we have mainly worked with two basic architectures, mostly adopting [32, 64, 96, 128] for simple datasets such as MNIST of Fashion MNIST, and using more complex structures such as [48, 96, 192, 384] for CIFAR10 or CelebA.
We also used different U-Net variants to extensively test the independence of the latent encoding discussed in Section~\ref{sec:gravitational}. 

\section{Embedding}
\label{sec:embedding}
We experimented with several different approaches for the embedding task. The most effective ones have been the direct synthesis through gradient descent, and the training of ad-hoc neural networks. Both techniques have interesting aspects worth discussing.

The gradient descent technique is intrinsically non-deterministic, producing a variegated set of  ``noisy" versions of a given image $x$, all able to reconstruct $x$ via reverse diffusion. 
The investigation of this set allows us to draw interesting conclusions on the shape of $emb(x)$. 

Gradient descent is, however, pretty slow. A direct network can be trained to compute a single element inside $emb(x)$. Interestingly enough, this single element seems to be very close to the {\em average} of all noisy versions of $x$ synthesized by the previous technique, suggesting evidence of its ``canonical" nature.  

The two techniques will be detailed in the following subsections.

\subsection{Gradient Descent Synthesis}
\label{sec:gradient_descent}
In Section~\ref{sec:gravitational}, we computed the shape of embeddings for a few synthetic datasets by defining a dense grid of points in the latent space and looking for their final mapping through the reverse denoising process. Unfortunately, the number of points composing the grid grows exponentially in the number of features, and the technique does not scale to more complex datasets.

\begin{figure}[ht]
\begin{center}
\begin{tabular}{cc}
\hspace{-.25cm}\includegraphics[width=.26\columnwidth]{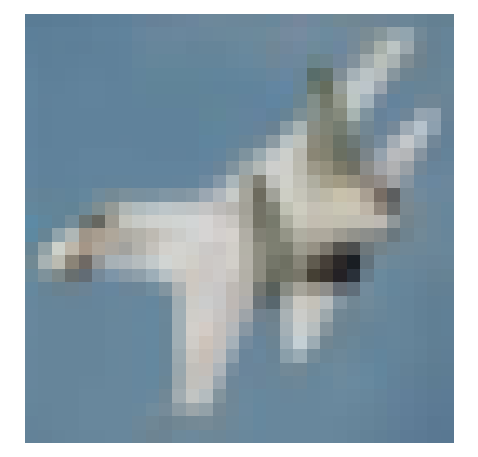}&\hspace{-.2cm}
\includegraphics[width=.63\columnwidth]{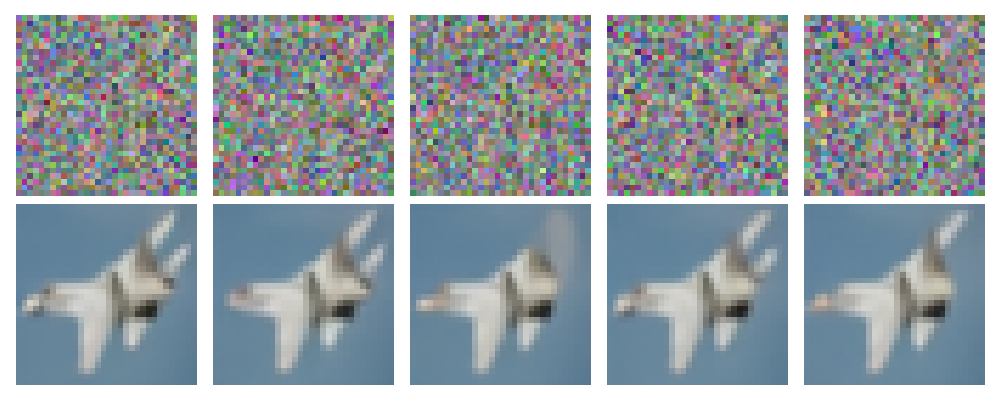}
\end{tabular}
\caption{Examples of seeds in the latent space. The image on the left is
the original.
On the right, we see 5 different seeds and their corresponding generations through the reverse diffusion process.}
\label{fig:example}
\end{center}
\end{figure}

A viable alternative is the gradient descent approach, where we synthesize inputs starting from random noise, using the distance from a given target image as the objective function. Generation usually requires several thousand steps, but it can be done in parallel on a batch of inputs. This allows us to compute, within a reasonable time, a sufficiently large number of samples in $emb(x)$ for any given $x$ (Figure~\ref{fig:example}). Having a full cloud of data, we can use standard techniques like PCA to investigate its shape, as well as to study how the image changes when moving along the components of the cloud (see Section~\ref{sec:PCA}). 


\begin{figure}[ht]
\begin{center}
\vspace{-.1cm}\begin{tabular}{cccc}
\hspace{-0.4cm}\includegraphics[width=.23\columnwidth]{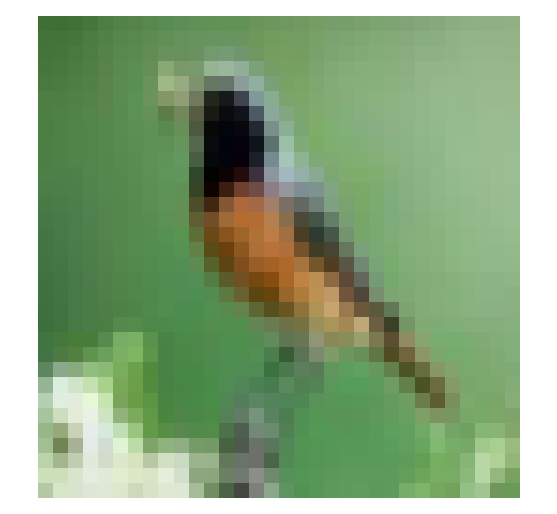}&
\hspace{-0.4cm}\includegraphics[width=.25\columnwidth]{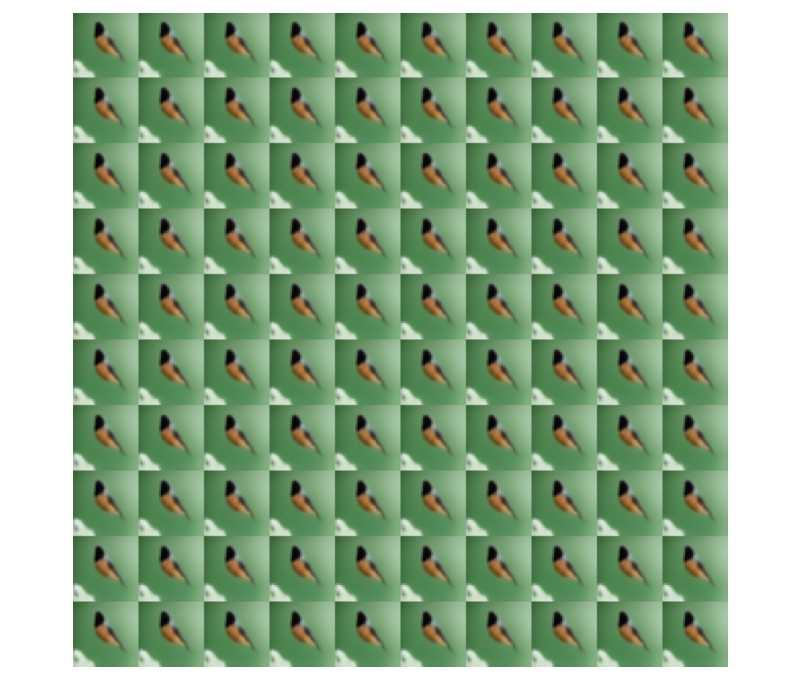} &
\hspace{-0.4cm}\includegraphics[width=.23\columnwidth]{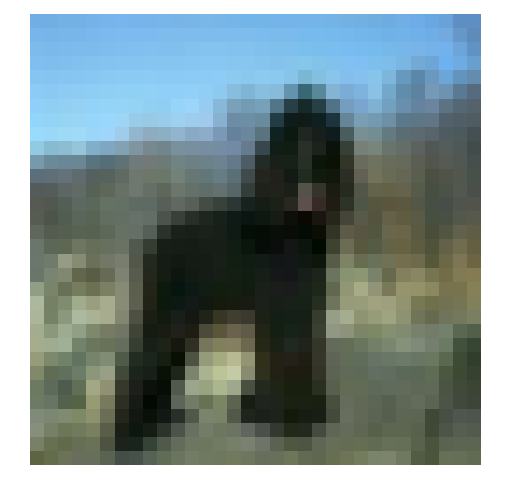}&
\hspace{-0.4cm}\includegraphics[width=.26\columnwidth]{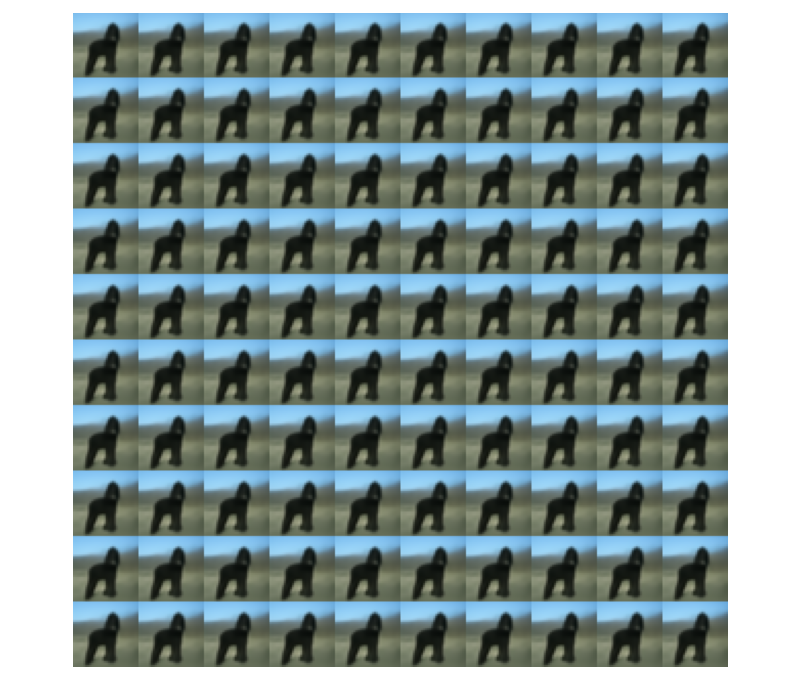}
\end{tabular}
\caption{Linear combination of seeds. Given the original image (1 and 3)
we compute by gradient descent a large cloud of seeds in its embedding. Then, we compute 100 {\em internal points}, as a 1-sum random linear combination of the given seeds.
Images 2 and 4 contain the results of these linear combinations. All generated images are similar between each other and are very close to the original image. Therefore, all internal points seem to belong to the embedding.
}
\label{fig:embedding1}
\end{center}
\end{figure}

A first interesting observation is that $emb(x)$ seems to be a convex region of the latent space. In Figure \ref{fig:embedding1} we show images obtained by reverse diffusion from 100 random {\em linear combinations} of seeds belonging to the embedding of the image on the left: all of them result in very similar reconstructions of the starting image. 

\begin{figure}[H]
\begin{center}
\includegraphics[width=.9\columnwidth]{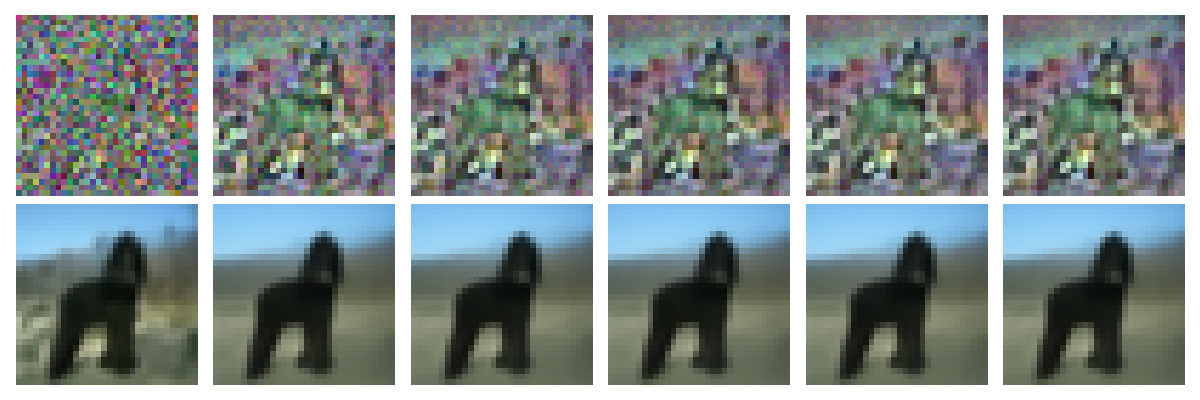}
\includegraphics[width=.9\columnwidth]{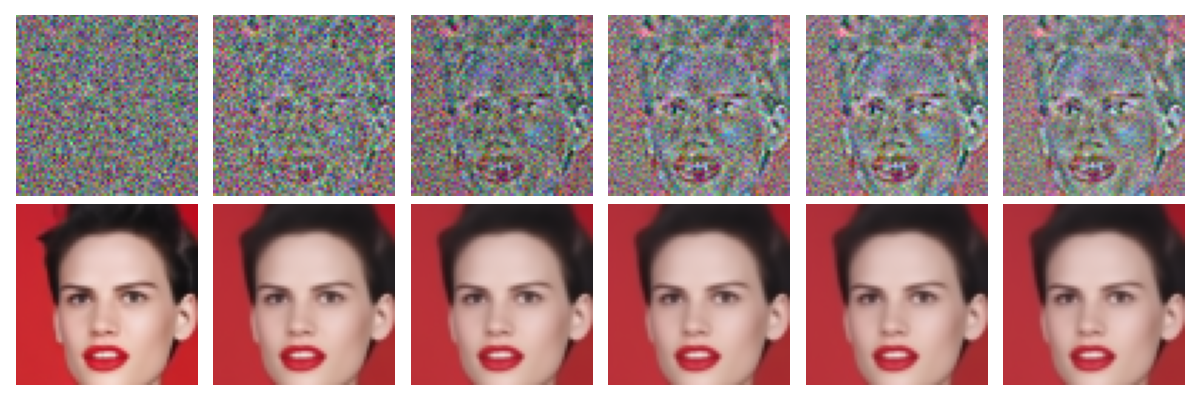}
\caption{Progressive averaging in CIFAR10 and CelebA. 
The first row shows seeds computed as the mean
of a progressive number of seeds in $emb(x)$; the second row shows their respective
output through the reverse denoising process. The output is very similar.
Additionally, observe that the original image becomes identifiable in the
mean.
}
\label{fig:mean}
\end{center}
\end{figure}

Due to the convexity of the space, its mean is also comprised in it. In Figure~\ref{fig:mean} we see the reconstructions obtained by considering as seed the average of a progressive number of seeds. The resulting images stabilize soon, although the result is slightly more blurry compared to using a single seed. The seeds on the borders of $emb(x)$ seem to provide slightly better reconstructions than internal points (which makes the quest for a ``canonical", high-quality seed even more challenging). 



\subsubsection{PCA Decomposition}
\label{sec:PCA}
Principal Component Analysis allows us to fit an ellipsoid over the cloud of datapoints, providing a major tool for investigating the actual shape of embeddings. 
According to the ``gravitational" intuition exposed in Section~\ref{sec:gravitational}, $emb(x)$ should be elongated along directions orthogonal to the data manifold: moving along those directions should not sensibly influence generation, which should instead be highly affected by movements along minor components. Moreover, since the data manifold is likely oriented along a relatively small number of directions (due to the low dimensionality of the manifold), we expect that most PCA components in each cloud will be orthogonal to the manifold, and have relatively high eigenvalues. 

\hspace{-.5cm}\begin{minipage}{5.5cm}
For instance, in the case of the clouds of seeds for CIFAR10, eigenvalues along all 3072 components typically span between 0.0001 and 4. We observe significant modifications only moving along the minor components of the clouds: in fact, they provide the shortest way to leave the embedding space of a given point.
However, as soon as we leave the embedding space of $x$ we should enter the embedding space of some ``adjacent" point $x^\prime$. In other words, the minor components should define directions {\em inside} the data manifold, and possibly have a ``semantical" (likely entangled) interpretation. 
\end{minipage}\hspace{.2cm}
\begin{minipage}{6cm}
\begin{figure}[H]
\begin{center}
\begin{tabular}{cc}
  \includegraphics[width=.45\columnwidth]{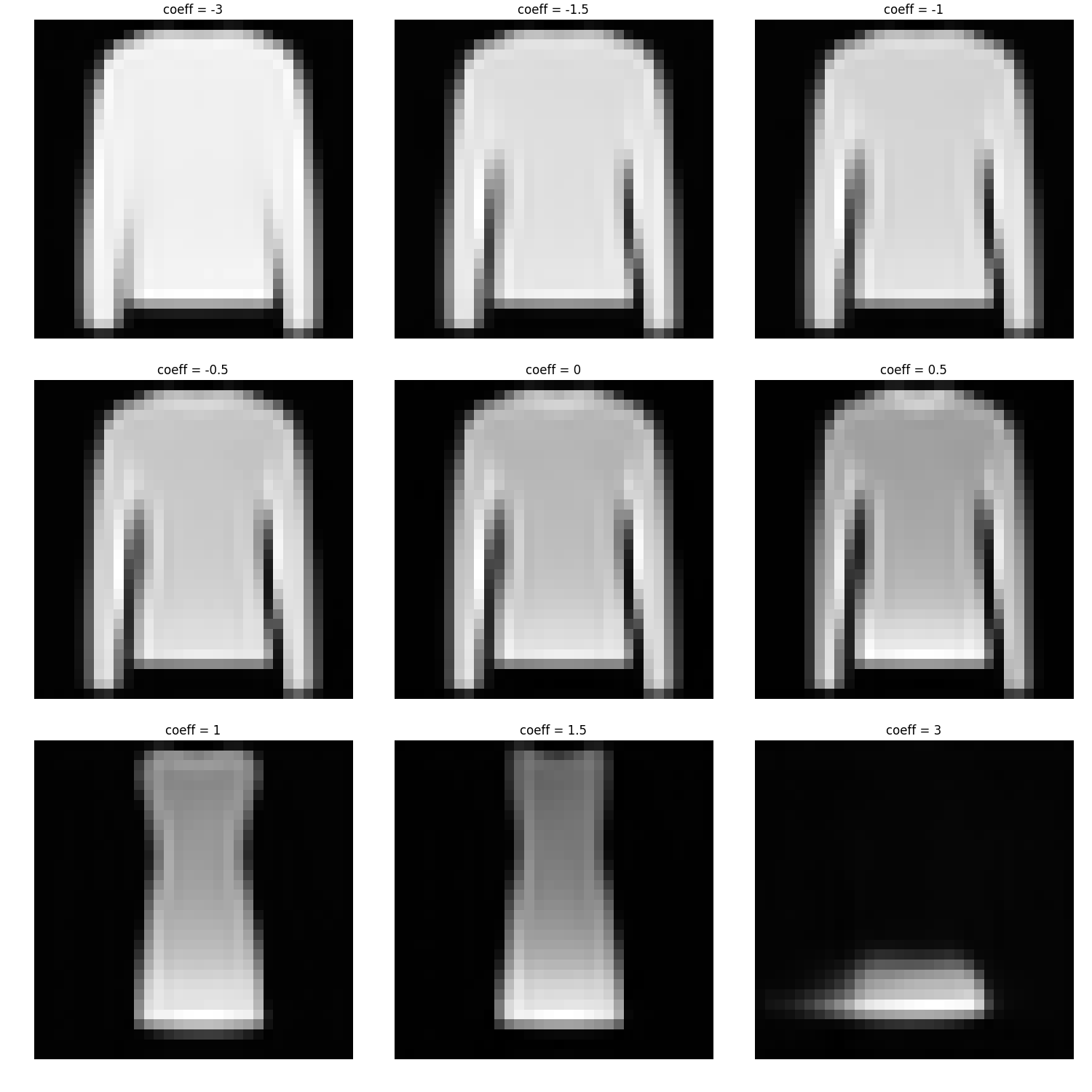}  &
  \includegraphics[width=.45\columnwidth]{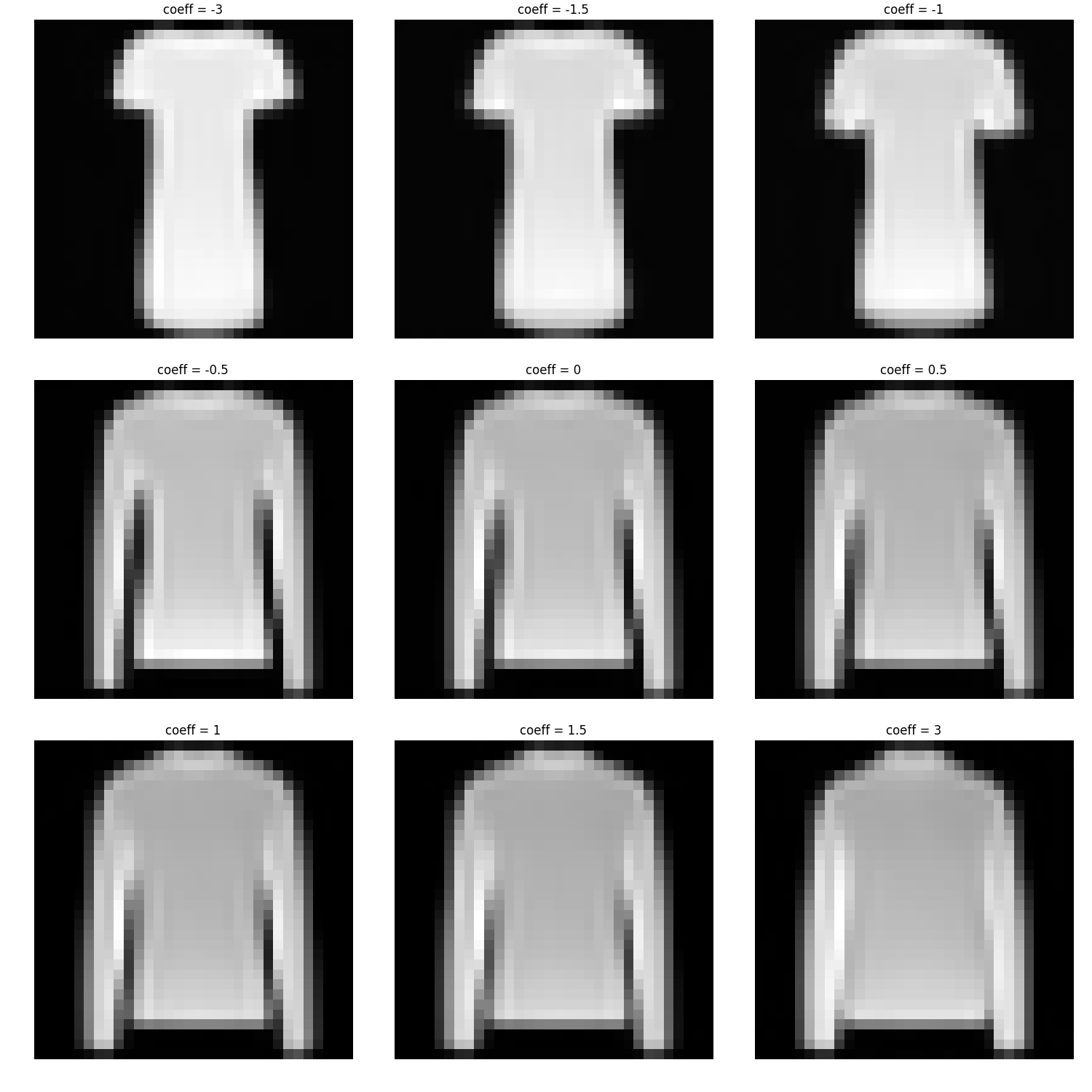} 
\end{tabular}
\caption{Fashion MNIST: movements along directions with minimal eigenvalues. 
The two groups of images refer to different components: starting from a mean seed
generating the image in the middle, we move along a given component by the indicated positive or negative factor of the normalized eigenvector. 
Observe the progressive change in intensity and shape. 
} 
\label{fig:fashion_mnist_minor_components}
\end{center}
\end{figure}
\end{minipage}


\subsection{Embedding Networks}
\label{sec:recoder}
The second approach consists in training a neural network to directly compute a sort of ``canonical" embedding for each image of the data manifold. The network takes as input an image $x$ and produces a seed $z_x \in emb(x)$; the loss function used to train the network is simply the distance between $x$ and the result $\hat{x}$ of the denoising process starting from $z_x$.

We tested several different networks; metrics relative to the most significant architectures are reported in Table~\ref{tab:embedding_networks}. 

\begin{table}[ht]
    \centering
    \begin{tabular}{c|c|c|c|c|c|c}
      {\bf Network} & {\bf Params} & \multicolumn{5}{c}{\bf MSE} \\
      & & MNIST & Fashion & CIFAR10 & Oxford & CelebA \\
      & &   &  MNIST & & Flowers  & \\\hline\hline 
      layers: 1 conv. $5\!\times\!5$ &  78 & .00704 & .0152 & .0303 & .0372 & .0189 \\\hline
      layers: 3 conv. $5\!\times\!5$  &  & & & & & \\
      channels: 16-16-out & 7,233 & .00271 & .00523 & .0090 & .0194 & .0101 \\\hline
      layers: 3 conv. $5\!\times\!5$ &  & & & &  \\
      channels: 64-64-out & 105,729 & .00206 & .00454 & .0061 & .0153 & .00829\\\hline
      layers: $\begin{array}{l}
               2 \text{conv.} 5\!\times\!5\\
               3 \text{conv.} 3\!\times\!3
            \end{array}$ & 859,009 & .00121 & .00172 & .0038 & .00882 & .00396  \\
      channels: & & & & & & \\     
      128-128-128-128-out & & & & & &\\\hline
      U-Net & 9,577,683 & .000361 & .000890 & .0012 & .00248 & .00147 \\\hline
    \end{tabular}
    \caption{Comparing the Mean Square Error (MSE) through the embedding-reconstruction process using different embedding networks; the MSE 
    standard deviation is below the last reported decimal. The number of parameters refers to the instance of the network for
    the CelebA dataset.} 
    \label{tab:embedding_networks}
\end{table}

A visual comparison of the behavior of the different networks is given in Figure~\ref{fig:different_networks_reconstruction}, relative to CIFAR10. More examples on CelebA are given below.

\begin{figure}[H]
\begin{center}
\includegraphics[width=\textwidth]{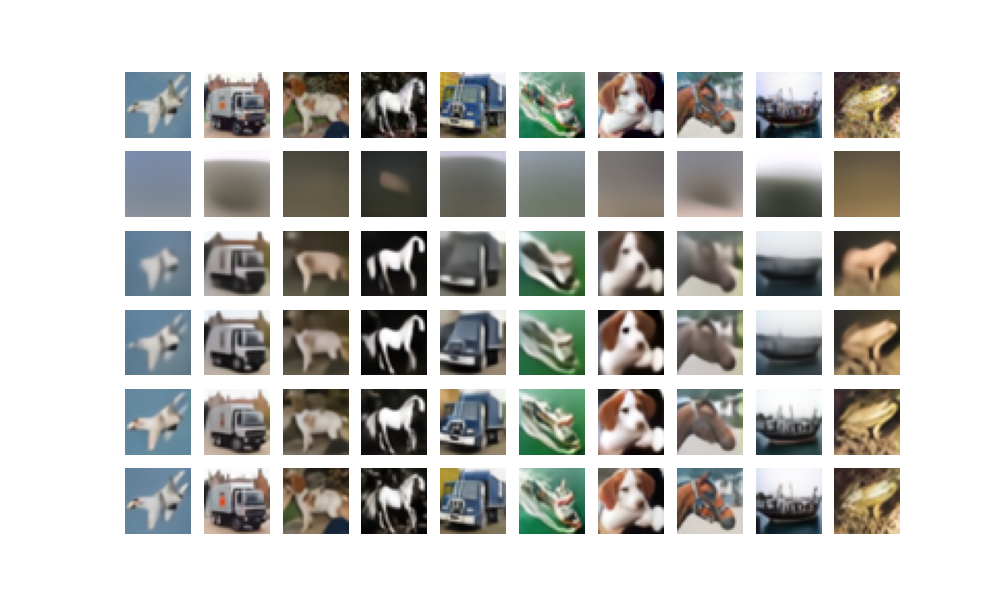}
\caption{Visual comparison with different Embedding Networks. We consider a set of test images from CIFAR10 (first row) and compute the embedding with one of the Embedding Networks of Table~\ref{tab:embedding_networks}. We then use the embeddings to generate the corresponding images (remaining rows). 
}
\label{fig:different_networks_reconstruction}
\end{center}
\end{figure}

We started our investigation with a very simple network: a single convolution with a $5\times 5$ kernel. The reason for this choice is that, according to the discussion we made in the introduction and the visualization of the mean element of the embedding clouds of Figure~\ref{fig:mean}, we expected the latent encoding to be similar to a suitably rescaled version of the source image. The results on a simple dataset like MNIST confirmed this hypothesis, but on more complex ones like CIFAR10 it does not seem to be the case, as exemplified in Figure~\ref{fig:different_networks_reconstruction}. We then progressively improved the model's architecture by augmenting their depth and channel dimensions, with the latter being typically the most effective way to improve their performance. In the end, the best results were obtained with a U-Net architecture that is practically identical to the denoising network. Many additional experiments have been performed, comprising autoencoders, residual networks, inception modules, and variants with different padding modalities or regularizations. However, they did not prove to be particularly effective and were thus dropped from our discussion.

\begin{figure}[ht]
\begin{center}
\includegraphics[width=\textwidth]{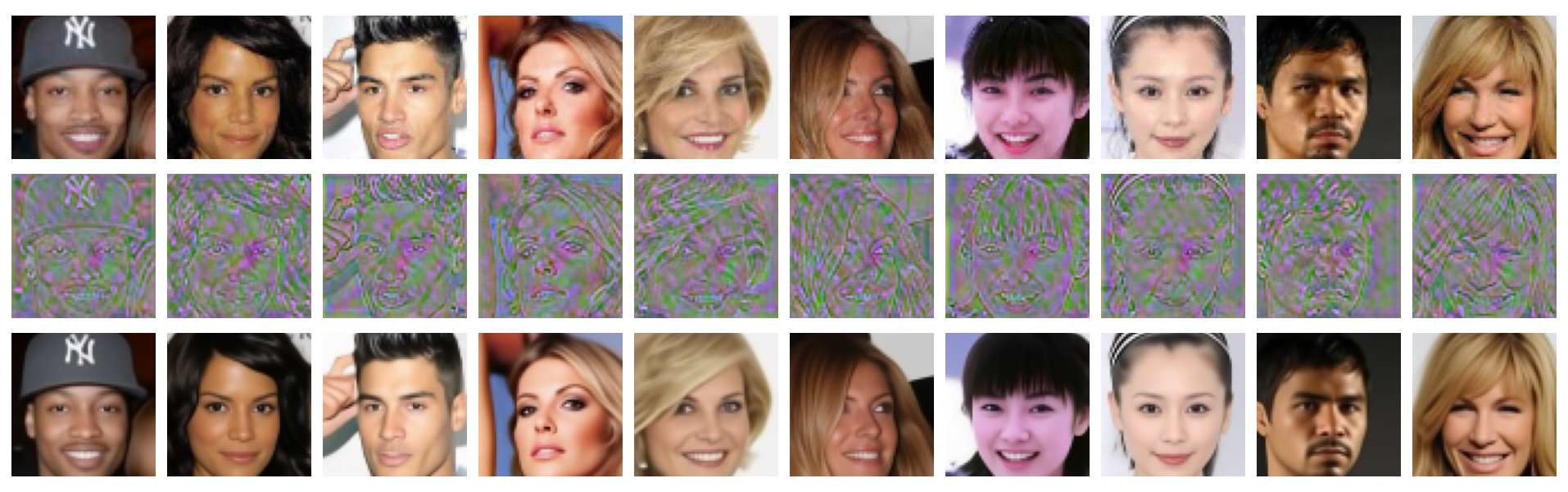}
\caption{Embedding examples for the CelebA dataset. The first row contains the original examples, the second the synthesized latent seed, and the third the reconstructed image. Reconstruction is very good, with just a slight blurriness.
}
\label{fig:celeba_gen_new}
\end{center}
\end{figure}

In Figure~\ref{fig:celeba_gen_new}, we show some examples of embeddings and relative reconstructions in the case of the CelebA dataset.

The quality of the reconstruction is definitely high, with just a slight blurriness. There are two possible justifications for the tiny inaccuracy of this result: it could either be a fault of the generator, which is unable to create the requested images (as it is frequently the case with Generative Adversarial Networks \cite{comparingNCAA}), or it could be a fault of the Embedding Network, which is unable to compute the correct seed.

To better investigate the issue, we performed two experiments. First, we restricted the reconstruction to images produced by the generator: in this case, if the Embedding network works well, it should be able to reconstruct almost perfect images. Secondly, we tried to improve the seeds computed by the Embedding Network through gradient descent, looking for better candidates. 

\begin{figure*}[ht]
\begin{center}
\includegraphics[width=\textwidth]{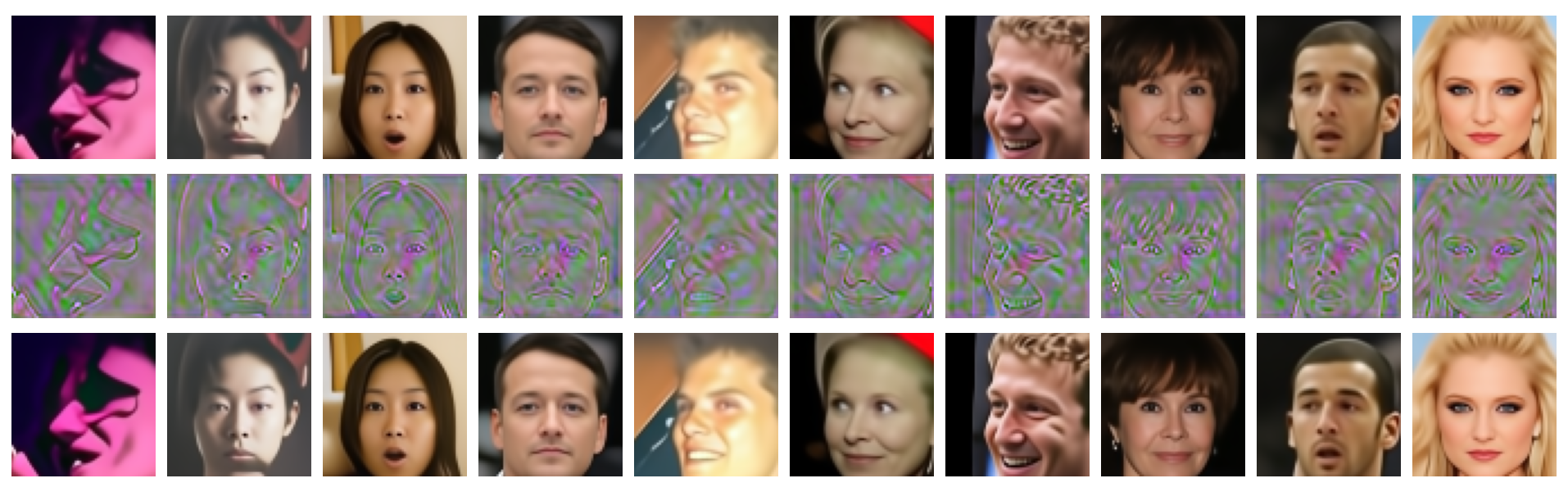}
\caption{Embedding examples on generated images. In this case, we start from
images created by the generator (first row) and re-embed them inside the latent space
(second row) using the Embedding Network. In the third row, we show the
reconstruction, which is almost perfect. This could be either explained by a deficiency of the generator, or just by the fact that generated images are ``simpler", and hence can be more easily embedded than real ones.
}
\label{fig:experiment1}
\end{center}
\end{figure*}

We report the result of the first experiment in Figure~\ref{fig:experiment1}.

While the reconstruction is qualitatively accurate, we can also confirm the effectiveness in a more analytical way. In Table~\ref{tab:mse for re-embedding} we compare the mean squared error of the reconstruction starting from original CelebA images versus generated data: the latter is sensibly smaller.
\begin{table}[h]
    \centering
    \begin{tabular}{c|c}
     {\bf Source Images}   & {\bf MSE} \\\hline\hline
    Dataset &  0.00147 \\\hline
    Generated    &  0.00074 \\\hline
    \end{tabular}
    \caption{Reconstruction error. In the first case, images are taken from the CelebA dataset: in the second case, they have been generated through the reverse diffusion process. The mean squared error (MSE) was computed over 1000 examples. Both experiments achieve a small reconstruction error, although the second one is even smaller.} 
    \label{tab:mse for re-embedding}
\end{table}

The fact that embedding works better for generated images is, however, not conclusive: it could either be explained by a deficiency of the generator, unable to generate all images in the CelebA dataset, or just by the fact that generated images are ``simpler" than real ones (observe the well-known patinated look, which is typical of most generative models) and hence more easily embeddable. 

\begin{figure}[ht]
\begin{center}
\includegraphics[width=\textwidth]{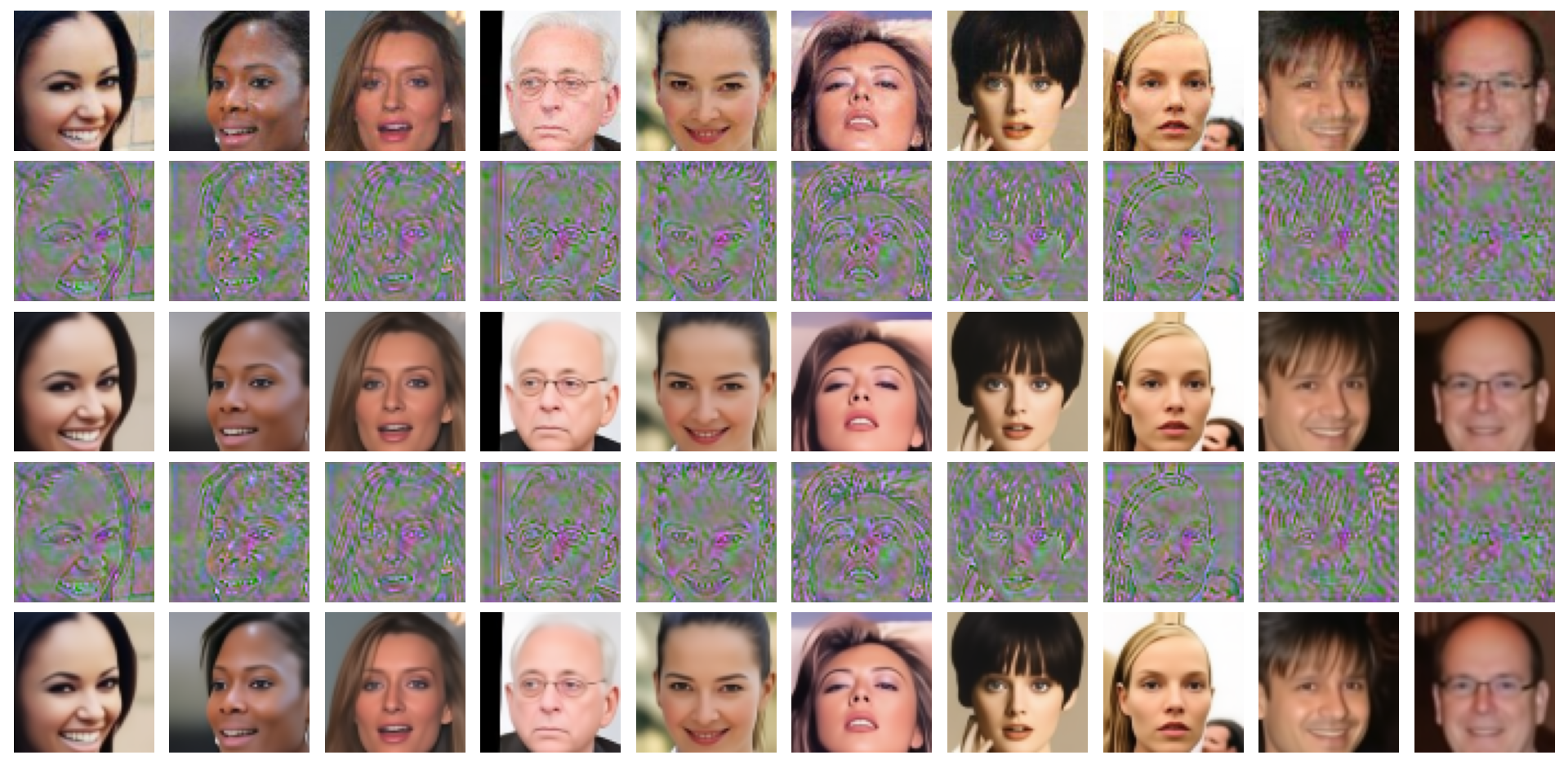}
\caption{Gradient descent fine-tuning. The seeds obtained through the embedding network (second row) are refined through gradient descent (fourth row). The respective resulting reconstructions are depicted in rows 3 and 5. The improvement is almost imperceptible.
}
\label{fig:experiment2}
\end{center}
\end{figure}

Even the results of the second experiment are not easily deciphered. 
From a visual point of view, refining the embedding through gradient descent is not producing remarkable results, as exemplified in Figure~\ref{fig:experiment2}.
However, numerically, we see an improvement from an MSE of $0.00147$ to an MSE of $0.00058$, which seems to suggest some margin of improvement for the embedding network.

In conclusion, both the generator and the embedder can likely still be improved.
However, a really interesting research direction seems to be the possibility to modify the latent representation to improve the realism of the resulting image, even if possibly not in the direction of the original. Therefore, a basic embedder, even if
not fully accurate, could still provide the starting point for very interesting 
manipulations.

\section{Conclusions}
\label{sec:conclusions}

In this article we addressed the problem of embedding data into the latent space of Deterministic Diffusion models, providing functionality similar to the encoder in a Variational Autoencoder, or the so-called \emph{recoder} for Generative Adversarial Networks. 
The main source of complexity when inverting a diffusion model is the non-injective nature of the generator: for each sample $x$, there exists a cloud of elements $z$ able to generate $x$. We call this set the embedding of $x$, denoted as $emb(x)$.
We performed a deep investigation of the typical shape of $emb(x)$, which suggests that embeddings are usually orthogonal to the dataset. %
These studies point to a sort of gravitational interpretation of the reverse diffusion process, according to which the space is progressively collapsing over the data manifold. In this perspective, $emb(x)$ is just the set of all trajectories in the space ending in $x$. 
We tested our interpretation on both low- and high-dimensional datasets, highlighting a quite amazing result: the latent space of a DDIM generator does not significantly depend on the specific generative model, but just on the data manifold.
In other words, passing the same seed as input to different DDIMs will result in almost identical outputs. 
In order to compute embeddings, we considered both gradient descent approaches, as well as the definition and training of specific Embedding Networks. We showed that, among all the architectures we tested, a U-Net obtained the best results, achieving a high-quality reconstruction from both a quantitative and qualitative point of view.

Embedding networks have a lot of interesting applications, largely exemplified in the introduction. More generally, the simplicity and ease of use of Embedding Networks open a wide range of fascinating perspectives about the exploration of semantic trajectories in the latent space, the disentanglement of the different aspects of variations, and the possibility of data editing. We thus hope that our results, by expanding the current understanding of generative models, can guide future research efforts.

\subsection*{Conflict of Interest} On behalf of all authors, the corresponding author states that there is no conflict of interest.

\bibliography{bibliography}

\end{document}